\pdfoutput=1
\documentclass{article}

    \PassOptionsToPackage{numbers, compress}{natbib}

\usepackage[final]{neurips_distshift_2023}

\usepackage{amsmath,amsfonts,bm}

\def\eqref#1{equation~\ref{#1}}

\def\1{\bm{1}}

\def\vx{{\bm{x}}}

\DeclareMathAlphabet{\mathsfit}{\encodingdefault}{\sfdefault}{m}{sl}
\SetMathAlphabet{\mathsfit}{bold}{\encodingdefault}{\sfdefault}{bx}{n}

\newcommand{\pdata}{p_{\rm{data}}}

\newcommand{\E}{\mathbb{E}}

\newcommand{\KL}{D_{\mathrm{KL}}}

\usepackage[utf8]{inputenc} %
\usepackage[T1]{fontenc}    %
\usepackage[colorlinks,linkcolor={blue!50!black},citecolor={blue!50!black},urlcolor={blue!50!black},backref=page,bookmarks=false]{hyperref} 
\usepackage{url}            %
\usepackage{booktabs}       %
\usepackage{amsfonts}       %
\usepackage{nicefrac}       %
\usepackage{microtype}      %
\usepackage{xcolor}         %

\usepackage{geometry}
\usepackage{multirow}
\usepackage{tabularx}
\usepackage{subcaption}

\usepackage{todonotes}

\usepackage[nameinlink]{cleveref}
\crefname{equation}{eq.}{eqs.}
\Crefname{equation}{Eq.}{Eqs.}
\usepackage{mathtools}
\usepackage{bm}
\usepackage{array} %
\usepackage{colortbl}
\definecolor{Gray}{gray}{0.94}
\definecolor{TuebingenRed}{rgb}{0.553, 0.176, 0.224} 
\definecolor{NiceRed}{rgb}{0.898, 0.490, 0.506}
\newcolumntype{a}{>{\columncolor{Gray}}c}
\newlength\savewidth

\newcommand{\dtrain}{\mathcal{D}_{\mathrm{train}}}
\newcommand{\dtest}{\mathcal{D}_{\mathrm{test}}}
\newcommand{\g}{\,|\,}

\newcommand{\fid}[2]{\mathrm{FID}(#1, #2)}
\newcommand{\is}[1]{\mathrm{IS}(#1)}

\newcommand{\norm}[1]{\lVert#1 \rVert}

\title{The SVHN Dataset Is Deceptive for Probabilistic Generative Models Due to a Distribution Mismatch}

\author{Tim Z. Xiao\textsuperscript{1,2,*} ~~~~~~Johannes Zenn\textsuperscript{1,2,*} ~~~~~~Robert Bamler\textsuperscript{1} \\
\textsuperscript{1}University of T\"ubingen ~~~\textsuperscript{2}IMPRS-IS \\ \footnotesize{\textsuperscript{*}Equal contribution, order determined by coin flip.}\\
\texttt{\{zhenzhong.xiao, johannes.zenn, robert.bamler\}@uni-tuebingen.de}
}

\begin{document}

\maketitle

\begin{abstract}

The Street View House Numbers (SVHN) dataset \citep{netzer2011reading} is a popular benchmark dataset in deep learning.
Originally designed for digit classification tasks, the SVHN dataset has been widely used as a benchmark for various other tasks including generative modeling.
However, with this work, we aim to warn the community about an issue of the SVHN dataset as a benchmark for generative modeling tasks: we discover that the official split into training set and test set of the SVHN dataset are not drawn from the same distribution.
We empirically show that this distribution mismatch has little impact on the classification task (which may explain why this issue has not been detected before), but it severely affects the evaluation of probabilistic generative models, such as Variational Autoencoders and diffusion models.
As a workaround, we propose to mix and re-split the official training and test set when SVHN is used for tasks other than classification.
We publish a new split and the indices we used to create it at \href{https://jzenn.github.io/svhn-remix/}{https://jzenn.github.io/svhn-remix/}.
\end{abstract}

\section{Introduction}

The Street View House Numbers (SVHN) dataset~\citep{netzer2011reading} is a popular benchmark datasets originated that from computer vision.
SVHN consists of real-world images from house numbers found on Google Street View and has 10 classes, one for each digit.
It is often treated as a more difficult variant of the MNIST dataset~\citep{lecun1998mnist}.
The dataset is divided into a training set $\mathcal{D}_{\mathrm{train}}$ with 73,257 samples, a test set $\mathcal{D}_{\mathrm{test}}$ with 26,032 samples, and a less used extra training set of 531,131 simpler samples.
In addition to classification tasks \citep{sermanet2012convolutional, hubara2016binarized, lin2013network, liu2017learning}, SVHN also serves as a benchmark for tasks such as generative modeling \citep{chen2016infogan, maaloe2016auxiliary}, out-of-distribution detection \citep{lee2018simple, serra2019input, xiao2020likelihood, nalisnick2018deep}, and adversarial robustness \citep{kos2018adversarial, song2018generative}.

As a toy dataset consisting of color images, SVHN is often used during the development of new generative models such as Generative Adversarial Networks (GANs;~\citep{goodfellow2020generative}), Variational Autoencoders (VAEs; \citep{kingma2013auto}), normalizing flows~\citep{rezende2015variational}, and diffusion models~\citep{song2021scorebased, ho2020denoising}.
To evaluate generative models, one commonly measures the sample quality using Fréchet Inception Distance (FID;~\citep{heusel2017gans}) and Inception Score (IS; \citep{salimans2016improved}).
In the following, we refer to likelihood-based generative models (like VAEs, normalizing flows, and diffusion models but not, e.g., GANs) as \emph{probabilistic} generative models as they explicitly model a probability distribution $p_\theta(\vx)$, where $\theta$ are the model parameters.
One typically also evaluates these models using (an approximation of) their likelihoods on test data.
This test set likelihood indicates how closely a model $p_\theta(\vx)$ approximates the true (usually inaccessible) data distribution $\pdata(\vx)$ from which data points are drawn.
It becomes particularly relevant if we want to use the model for tasks such as out-of-distribution detection and lossless data compression.

Surprisingly, we discover that SVHN, as a popular benchmark, has a \emph{distribution mismatch} between its training set $\mathcal{D}_{\mathrm{train}}$ and its test set $\mathcal{D}_{\mathrm{test}}$.
In other words, $\mathcal{D}_{\mathrm{train}}$ and $\mathcal{D}_{\mathrm{test}}$ do not seem to come from the same distribution.
We find that this mismatch has little effect on classification tasks for supervised learning, or on sample quality for generative modeling.
However, we show that for probabilistic generative models such as VAEs and diffusion models, the mismatch leads to a false assessment of model performance when evaluating test set likelihoods:
test set likelihoods on the SVHN dataset are deceptive since $\mathcal{D}_{\mathrm{test}}$ appears to be drawn from a simpler distribution than~$\mathcal{D}_{\mathrm{train}}$.
As a workaround, we merge the original $\mathcal{D}_{\mathrm{train}}$ and $\mathcal{D}_{\mathrm{test}}$, then shuffle and re-split them.
We empirically show that this remixing solves the problem of distribution mismatch, thus restoring the SVHN test set likelihood as an informative metric for probabilistic generative models.
We also publish the new split we used in our experiments as a proposal of a canonical split for future research on generative models.

\section{Distribution Mismatch in SVHN}

In this section we show evidence for the distribution mismatch in SVHN between the training set $\dtrain$ and the test set $\dtest$.
We downloaded the SVHN dataset from its official website%
\footnote{\href{http://ufldl.stanford.edu/housenumbers/}{http://ufldl.stanford.edu/housenumbers/}}, which is also the default download address used by \texttt{Torchvision} \citep{torchvision2016} and \texttt{TensorFlow Datasets} \citep{TFDS}.
\begin{figure}[t]
    \centering
    \includegraphics[width=0.9\textwidth]{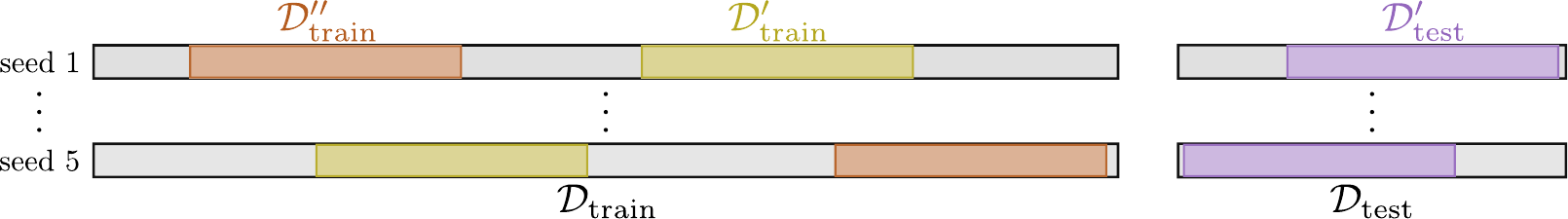}
    \caption{
    Five random splits (with reshuffle) of $\mathcal{D}_\mathrm{train}$ and $\mathcal{D}_\mathrm{test}$ into $\dtrain'$, $\dtrain''$, and $\dtest'$.
    }
    \label{fig:train-test-split}
\end{figure}

\subsection{Defining Distribution Mismatch} \label{sec:distribution-mismatch}

We often assume $\dtrain$ and $\dtest$ in a benchmark dataset consists of i.i.d.~samples from an underlying data distribution $\pdata(\vx)$.
Thus, given a distance metric $D(p_1(\vx) \, , p_2(\vx))$ that measures the dissimilarity between two distributions $p_1(\vx)$ and $p_2(\vx)$, we expect that
\begin{align}\label{eq:distance-to-pdata}
  D\big(\pdata(\vx), \dtrain' \big) \approx D\big(\pdata(\vx), \dtest' \big)
\end{align}
and that both sides have a low value.
Here, $\dtrain'$ and $\dtest'$ are equally sized random subsets of $\dtrain$ and~$\dtest$, respectively (this will simplify comparisons both within and across datasets below).
In practice, we typically do not have access to~$\pdata(\vx)$.
But we can bypass this issue by drawing an additional random subset $\dtrain''$ of $\dtrain$, which has the same size as $\dtrain'$ and does not overlap with it (see \Cref{fig:train-test-split}).
Then, \Cref{eq:distance-to-pdata} also holds if we replace $\dtrain'$ with~$\dtrain''$, and by combining the two variants of \Cref{eq:distance-to-pdata} for $\dtrain'$ and~$\dtrain''$, we find by the triangle inequality,
\begin{align} \label{eq:sim}
  D\big(\dtrain'', \dtrain'\big) \approx D\big(\dtrain', \dtest'\big)
\end{align}
and that, again, both sides should have a low value.
Conversely, if $D(\dtrain'', \dtrain')$ differs substantially from $D(\dtrain'', \dtest')$, it indicates a distribution mismatch.
Note that \Cref{eq:sim} is a necessary but not a sufficient condition for matching distributions.

\subsection{Evaluation Using Sample Quality Measures} \label{sec:evaluation-vision-metrics}

To detect a distribution mismatch between any two sets of images, we need to find a well-motivated distance metric $D(\,\cdot \, , \cdot\,)$ to use in \Cref{eq:sim}.
Here, we draw inspiration from the Fréchet Inception Distance (FID)~\citep{heusel2017gans}, which is usually used to measure sample quality, but which we repurpose for our setup.
We contrast FID to Inception Score (IS)~\citep{salimans2018improving}, which does \emph{not} compare two sets of images.

\paragraph{Fréchet Inception Distance (FID)}
measures semantic dissimilarity between two finite sets $\mathcal D_1$ and~$\mathcal D_2$ of images.
One first maps both sets into a semantic feature space using a feature extractor~$f(\cdot)$.
Then, one computes the feature means $\bm{\mu}_1, \bm{\mu}_2$ and covariances $\bm{\Sigma}_1, \bm{\Sigma}_2$, which parameterize two Gaussian distributions.
FID is defined as the Fréchet distance between these two Gaussians,
\begin{align}
    \label{eq:fid}
        \fid{\mathcal D_1}{\mathcal D_2}
        =
        {\norm{\bm{\mu}_1 - \bm{\mu}_2}}_2^2
        + 
        \mathrm{Tr}
        \left(
            \bm{\Sigma}_1 + 
            \bm{\Sigma}_2 -
            2 \left(\bm{\Sigma}_1 \bm{\Sigma}_2 \right)^{1/2}
        \right).
\end{align}
For image data, $f(\cdot)$ is most commonly the activation at the penultimate layer of an Inception classifier~\citep{szegedy2015going} that was pre-trained on ImageNet.
A lower FID indicates higher semantic similarity.

FID is commonly used to evaluate sample quality by comparing samples $\vx\sim p_\theta(\vx)$ from a trained generative model to~$\dtrain$.
We instead apply FID directly as the distance metric~$D$ on both sides of \Cref{eq:sim}, without training any generative model.
Thus, we randomly sample (without replacement) three subsets $\dtrain'$, $\dtrain''$, and $\dtest'$ from $\dtrain$ and $\dtest$, respectively, each of size ${10,000}$.
Then, we assess \Cref{eq:sim} by calculating $\fid{\dtrain''}{\dtrain'}$ and $\fid{\dtrain''}{\dtest'}$.
We repeat this procedure over $5$ different random seeds as illustrated in \Cref{fig:train-test-split} and report FID means and standard deviations in \Cref{tab:fid_is_all}.
For comparison, we also evaluate CIFAR-10 \citep{krizhevsky2009learning} using the same procedure.

\Cref{tab:fid_is_all} shows that $\fid{\dtrain''}{\dtrain'}$ differs significantly from $\fid{\dtrain''}{\dtest'}$, violating the necessary condition \Cref{eq:sim}, therefore indicating that there is a distribution mismatch between $\dtrain$ and $\dtest$ for SVHN.
As a comparison, for CIFAR-10, the two FIDs are basically indistinguishable.

\begin{table}[t]
    \renewcommand{\arraystretch}{1.2}
    \centering
    \small
    \caption{
    FID (lower means larger similarity) and IS (higher means better sample quality) on three datasets, averaged over $5$ random seeds.
    For SVHN, we find that the FID between random subsets of the training and test set (bold red) is significantly higher than the FID between non-overlapping subsets of the training set of the same size, while 
    IS for $\dtrain'$ and $\dtest'$ is similar within all datasets.
    }
    \vspace{0.5em}
    \begin{tabularx}{\linewidth}{ l | 
                                >{\centering\arraybackslash}X 
                                >{\centering\arraybackslash}X 
                                >{\centering\arraybackslash}X }
    $\mathrm{FID}\; (\downarrow), \; \mathrm{IS}\; (\uparrow)$  & SVHN & \cellcolor{Gray}{SVHN-Remix} & CIFAR-10\\
    \specialrule{1pt}{1pt}{1pt} %
    $\fid{\dtrain''}{\dtrain'}$ 
    & $3.309 \pm 0.029$ 
    & \cellcolor{Gray}{$3.334\pm 0.018$} 
    & $5.196\pm 0.040$\\
    $\fid{\dtrain''}{\dtest'}$ 
    & \textcolor{TuebingenRed}{$\bm{16.687}\pm\bm{0.325}$\hspace{0.5em}} 
    & \cellcolor{Gray}{$3.326\pm 0.015$} 
    & $5.206\pm 0.031$ \\
    \hline
    $\is{\dtrain' \,|\, \bar{\mathcal D}_\mathrm{train}}$ 
    & $8.507\pm 0.114$ 
    & \cellcolor{Gray}{$8.348\pm 0.568$} 
    & $7.700\pm 0.043$ \\
    $\is{\dtest' \;\;\! \,|\, \bar{\mathcal D}_\mathrm{train}}$ 
    & $8.142\pm 0.501$ 
    & \cellcolor{Gray}{$8.269\pm 0.549$} 
    & $7.692\pm 0.023$ \\
    \end{tabularx}
    \label{tab:fid_is_all}
\end{table}

\paragraph{Inception Score (IS)}
is another common measure of sample quality.
Unlike FID, IS does not build on a similarity metric between samples.
Instead, it evaluates how well data points in a set~$\mathcal D$ can be classified with a classifier $p_\mathrm{cls.}(y \g \vx)$ that was trained on~$\dtrain$, and how diverse their labels~$y$ are,
\begin{align}
    \label{eq:is}
    \is{\mathcal{D} \,|\, \dtrain} 
    =
    \exp
    \big(
    \E_{\vx \sim \mathcal{D}}
    \big[\KL\left[
        p_\mathrm{cls.}(y \g \vx)
        \;\|\;
        p_\mathrm{cls.}(y)
    \right]\big]
    \big),
\end{align}
where $\KL$ denotes the Kullback-Leibler divergence \citep{kullback1951information}, and $p_\mathrm{cls.}(y) = \E_{ \vx \sim \mathcal{D}} [ p_\mathrm{cls.}(y \g \vx) ]$.

Similar to FID discussed above, the generative modeling literature typically applies IS to samples $\vx\sim p_\theta(\vx)$ from a trained generative model.
We instead apply it directly to subsets of the SVHN dataset to measure their quality.
We randomly sample (without replacement) subsets $\dtrain'$ and $\dtest'$ of $\dtrain$ and $\dtest$, respectively, each with size $M=10,000$.
We then train a classifier $p_\mathrm{cls.}(y \g \vx)$ on $\bar{\mathcal D}_\mathrm{train} := \dtrain\backslash\dtrain'$, and we evaluate the IS on both $\dtrain'$ and $\dtest'$.
See \Cref{app:sec:details-classifier} for the model architecture of $p_\mathrm{cls.}(y \g \vx)$.
We follow the same procedure for CIFAR-10.

The results in \Cref{tab:fid_is_all} show that there is not much difference between $\is{\dtrain' \,|\, \bar{\mathcal D}_\mathrm{train}}$ and $\is{\dtest' \,|\, \bar{\mathcal D}_\mathrm{train}}$ for both SVHN and CIFAR-10.
Thus, in terms of the class distribution $p(y)$ and the data quality, $\dtrain$ and $\dtest$ are similar for both SVHN and CIFAR-10, which is expected for classification benchmark datasets.
It also tells us that if we want to measure the sample quality in terms of distribution similarity, we should not use IS as the metric.

\section{SVHN-Remix}

As a workaround to alleviate distribution mismatch in SVHN, we propose a new split called SVHN-Remix, which we created by joining the original $\dtrain$ and $\dtest$, random shuffling, and re-splitting them into $\mathcal{D}_\mathrm{train}^\mathrm{remix}$ and $\mathcal{D}_\mathrm{test}^\mathrm{remix}$.
We make sure the size of the new training and test set is the same as before, and the number of samples for each class is also preserved for both the new training and test set. 
We evaluated SVHN-Remix using the same procedures as in Section~\ref{sec:evaluation-vision-metrics} using FID and IS. 
The results in \Cref{tab:fid_is_all} show that $\fid{{\dtrain^{\mathrm{remix}}}''}{{\dtrain^{\mathrm{remix}}}'}$ is now very similar to $\fid{{\dtrain^{\mathrm{remix}}}''}{{\dtest^{\mathrm{remix}}}'}$, i.e., just like CIFAR-10 it now satisfies \Cref{eq:sim}.
The result also shows that our remixing does not impact the IS, which further indicates that IS cannot be used for detecting distribution mismatch.

\section{Implications on Supervised Learning} \label{sec:implications-classifiers}

\begin{figure}
\centering
\begin{minipage}[t]{.242\textwidth}
    \centering
    \includegraphics[width=\linewidth]{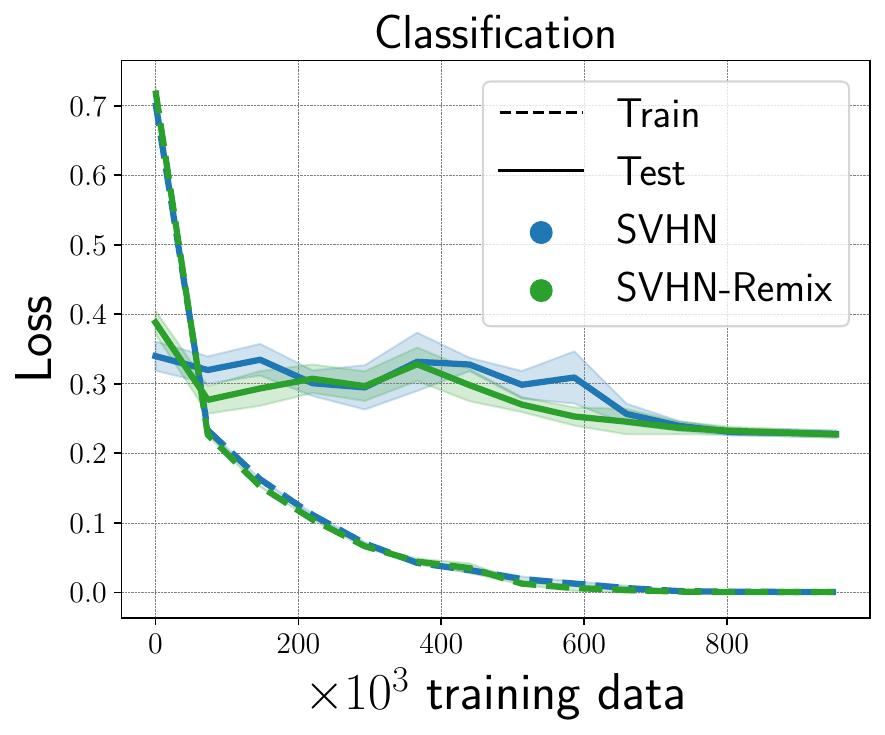}
    \subcaption{Classification}
    \label{fig:classifier}
\end{minipage}
\begin{minipage}[t]{.244\textwidth}
    \centering
    \includegraphics[width=\linewidth]{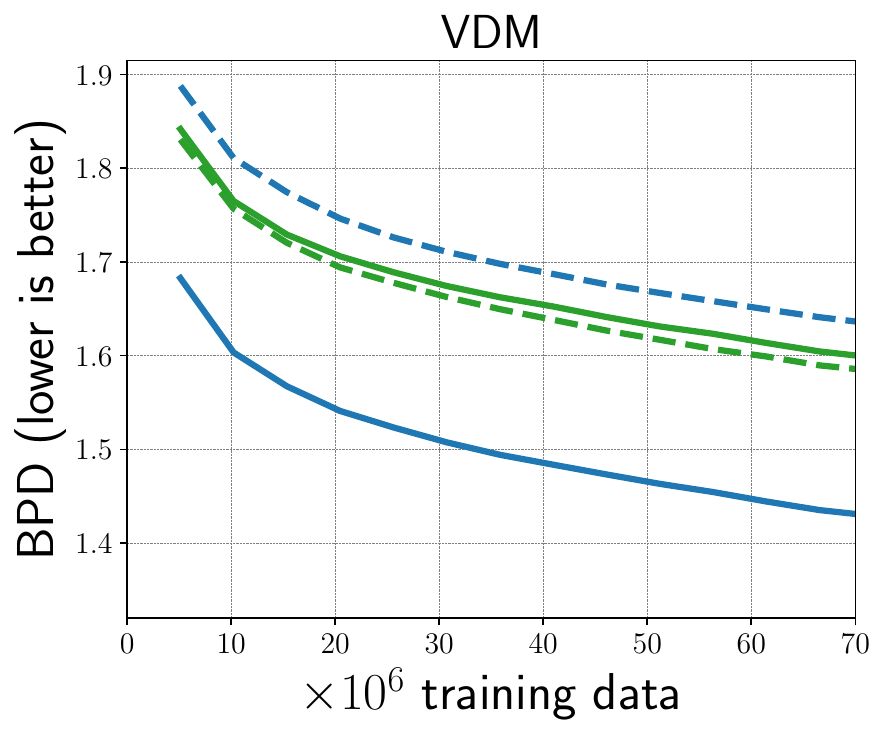}
    \subcaption{VDM}
    \label{fig:bpd-vdm}
\end{minipage}%
\begin{minipage}[t]{.49\textwidth}
    \centering
    \includegraphics[width=\linewidth]{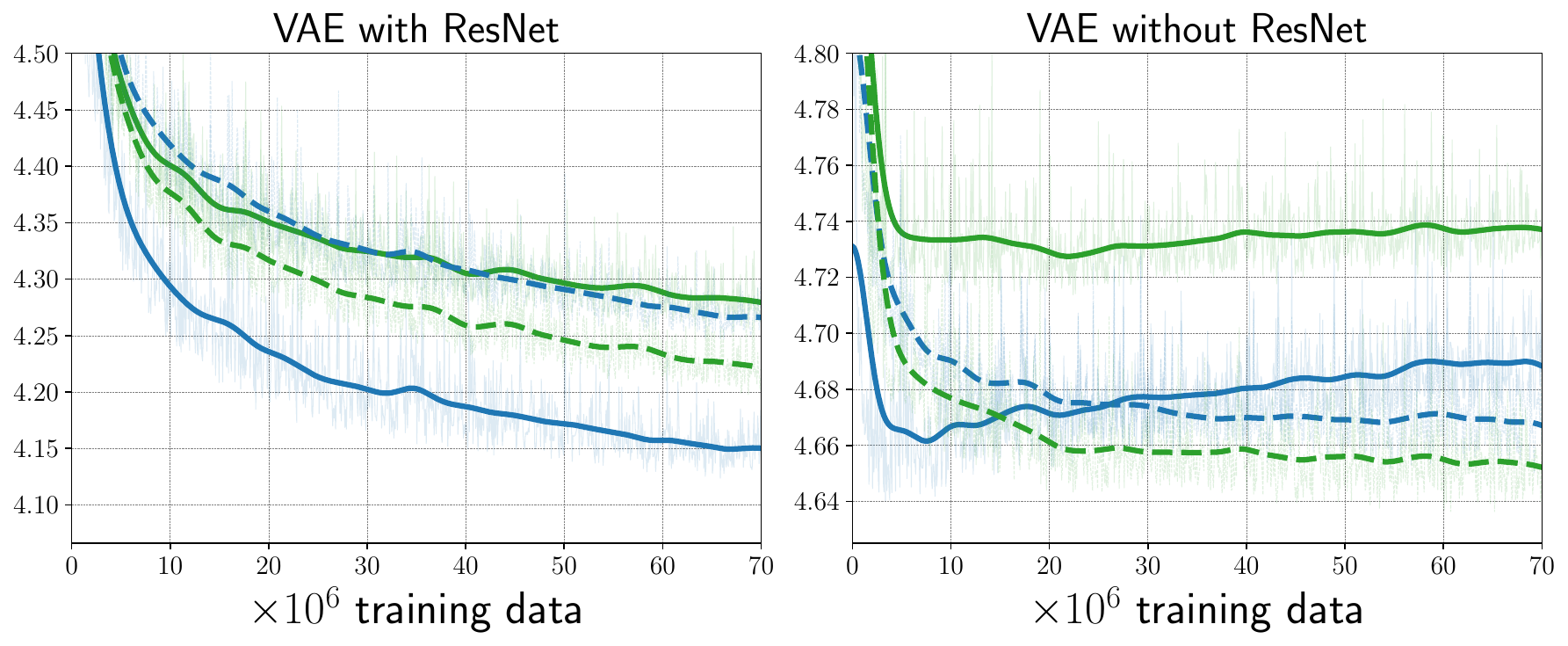}
    \subcaption{VAEs}
    \label{fig:bpd-vae}
\end{minipage}
\caption{
(a): 
classification loss evaluated on training set (dashed) and test set (solid) on SVHN (blue) and SVHN-Remix (green) for five random seeds (lines are means, shaded areas are~$\pm\sigma$). 
The losses are similar (details in \Cref{sec:implications-classifiers}).
(b) and (c): BPD evaluated as a function of training progress on the training set (dotted) and test set (solid) for SVHN (blue) and SVHN-Remix (green).
For SVHN, the order of training and test set performance is flipped compared to SVHN-Remix (details in \Cref{sec:implications-generative-models}).
}
\label{fig:bpd-classifier}
\end{figure}

In the classification setting, we want to learn a conditional distribution $p_\mathrm{cls.}(y \g \vx)$ by maximizing the cross entropy $\E_{(\vx, y) \sim \dtrain}[p_\mathrm{cls.}(y \g \vx)]$.
We train classifiers on the training sets of both the original SVHN and of SVHN-Remix (for classifier details see \Cref{app:sec:details-classifier}).
The results in \Cref{fig:classifier} show that the distribution mismatch in SVHN between $\dtrain$ and $\dtest$ has minimal impact on the classification loss.
This observation is consistent with the IS (which also relies on a classifier), where $\is{\dtrain'}$ and $\is{\dtest'}$ are also not affected by the proposed remixing.
Therefore, we suspect that the distribution mismatch in SVHN happens in-distribution, i.e., $\dtest$ covers only a subset of $\dtrain$ but more densely (we further verify this in \Cref{sec:implications-generative-models}).
Note other conceivable forms of distribution mismatch, such as $\dtest$ being out-of-distribution for $\dtrain$, would seriously impact the classification performance of $p_\mathrm{cls.}(y \g \vx)$ on $\dtest$.
The results in \Cref{fig:classifier} explain why the issue of distribution mismatch in SVHN has not been identified earlier, as SVHN is mostly used for classification.

\section{Implications on Probabilistic Generative Models} \label{sec:implications-generative-models}

Probabilistic generative models $p_\theta(\vx)$, such as Variational Autoencoders (VAEs; \citep{kingma2013auto,rezende2015variational}) and diffusion models \citep{sohl2015deep,song2021scorebased}, aim to explicitly model the underlying data distribution $\pdata(\vx)$.
These probabilistic models are the backbone of foundation models.
Developing and fast prototyping such models often involves using small benchmark datasets like SVHN.
We normally evaluate these models by their likelihood on the test set, i.e., $p_\theta(\dtest)$.
Hence, a distribution mismatch between $\dtrain$ and $\dtest$ can lead to false evaluation of these models.

In order to see the impact of distribution mismatch on probabilistic generative models, we train and evaluate a variational diffusion model (VDM; \citep{kingma2021variational}) and two VAEs (one with ResNet \citep{ho2020denoising} architecture, one without) on both SVHN and SVHN-Remix (details see \Cref{app:sec:details-pgm}).
We report model performance in bits per dimension (BPD), which is proportional to negative log-likelihood of the model on a dataset.
Lower BPD corresponds to higher likelihood.
\Cref{fig:bpd-vdm} and \subref{fig:bpd-vae}(left) show that both VDM and VAE with ResNet have much lower BPD on $\dtest$ than on $\dtrain$ (solid blue) when trained with SVHN.
\emph{This means that both models have higher likelihood on $\dtest$, even though they are trained to maximize the likelihood on $\dtrain$.}
This is exactly due to the distribution mismatch, where $\dtest$ is in-distribution for $\dtrain$ but has much higher density on the `easy' data.
Otherwise, we would expect a slightly lower BPD on $\dtrain$ than on $\dtest$, as we indeed observe when using SVHN-Remix (green lines).
Additionally, \Cref{fig:bpd-vae}(right) shows the results for the VAE without ResNet, where the solid blue line first goes below the dashed blue line, then goes above it.
This occurs because the VAE begins to overfit $\dtrain$, which does not invalidate our previous findings.
In summary, when there is distribution mismatch between $\dtrain$ and $\dtest$, the likelihood evaluated on $\dtest$ can be misleading and does not provide meaningful information on the generalization performance of the model.

\section{Conclusion}

In this paper, we show that there is a distribution mismatch between the training set and test set in the SVHN dataset.
This distribution mismatch affects the evaluation of probabilistic generative models such as VAEs and diffusion models, but does not harm classification. 
We provide a new split of the SVHN dataset resolving this issue.
In a broader sense, this tells us that when creating benchmark datasets for (probabilistic) foundation models we have to be mindful of a distribution mismatch.

\newpage
\subsubsection*{Acknowledgments}
The authors would like to thank Takeru Miyato, Yingzhen Li, and Andi Zhang for helpful discussions.
Funded by the Deutsche Forschungsgemeinschaft (DFG, German Research Foundation) under Germany’s Excellence Strategy~--~EXC number 2064/1~--~Project number 390727645.
This work was supported by the German Federal Ministry of Education and Research (BMBF): Tübingen AI Center, FKZ:~01IS18039A.
Robert Bamler acknowledges funding by the German Research Foundation (DFG) for project 448588364 of the Emmy Noether Programme.
The authors thank the International Max Planck Research School for Intelligent Systems (IMPRS-IS) for supporting Tim Z.~Xiao and Johannes Zenn.

\paragraph{Reproducibility Statement.}
We publish the new split and the indices we used to create it at \href{https://jzenn.github.io/svhn-remix/}{https://jzenn.github.io/svhn-remix/}.

\bibliographystyle{plain}
\bibliography{ref}

\newpage
\appendix

\section{Details on the Probabilistic Generative Models} \label{app:sec:details-pgm}

\subsection{VAEs} \label{app:sec:details-vae}
The VAEs in this work use standard Gaussian priors and diagonal Gaussian distributions for the inference network.
Our VAEs are trained with two architectures: a residual architecture \citep{he2016deep} and a non-residual architecture.
The generative model uses a a discretized mixture of logistics (MoL) likelihood \citep{salimans2017pixelcnn}.

The VAE with residual architecture has two convolutional layers (kernel size: $4$, stride: $2$, padding:~$1$), a residual layer, and a final convolutional layer.
The resulting latent has a dimension of $64$.
The residual layer sequentially processes the input by two convolutional layers (kernel size: $3$, stride: $1$, padding: $1$ and kernel size: $1$, stride: $1$, padding: $0$).
For all convolutional layers we use BatchNorm \citep{ioffe2015batch}.
The decoder mirrors the architecture of the encoder.

The non-residual architecture passes the input through three convolutional layers ((i) kernel size: $3$, stride: $2$, padding: $1$; (ii) kernel size: $4$, stride: $2$, padding: $1$; (iii) kernel size: $5$, stride: $2$, padding: $1$) and maps the flattened output with a fully-connected layer to mean and variance, respectively.
The latent dimension is $20$.
The decoder mirrors the architecture of the encoder (three convolutional layers with (i) \& (ii) kernel size: $6$, stride: $2$, padding: $2$; (iii) kernel size: $5$, stride: $2$, padding: $1$) but uses transposed convolutions \citep{zeiler2010deconvolutional}.

\subsection{Diffusion Model} \label{app:sec:details-dm}
We use an open source implementation\footnote{\href{https://github.com/addtt/variational-diffusion-models}{https://github.com/addtt/variational-diffusion-models}} of Variational Diffusion Models \citep{kingma2021variational}.
We train two diffusion models (on SVHN and SVHN-Remix), each on 4 NVIDIA A100 40GB GPUs with a batch size of $512$ for approximately $2$ days.

\section{Details on the Classifier} \label{app:sec:details-classifier}

We train a classifier to compute the IS (see \Cref{sec:evaluation-vision-metrics}) and we train a classifier to compare training and test losses in \Cref{sec:implications-classifiers}.

We use a ResNet-18 \citep{he2016deep} for the classifiers trained on SVHN, and we use a DenseNet-121 \citep{huang2017densely} for classifiers trained on CIFAR-10.
Each classifier is trained for $14$ epochs with a batch size of $256$ on the corresponding dataset.
We use stochastic gradient descent \citep{robbins1951stochastic} with a learning rate of $0.001$, a momentum \citep{rumelhart1986learning} of $0.9$, and weight decay of $5\cdot10^{-4}$.

\end{document}